# Vector Quantization for Machine Vision

Vincenzo Liguori    Ocean Logic Pty Ltd, Australia    Email: enzo@ocean-logic.com

*Abstract*—This paper shows how to reduce the computational cost for a variety of common machine vision tasks by operating directly in the compressed domain, particularly in the context of hardware acceleration. Pyramid Vector Quantization (PVQ) is the compression technique of choice and its properties are exploited to simplify Support Vector Machines (SVM), Convolutional Neural Networks(CNNs), Histogram of Oriented Gradients (HOG) features, interest points matching and other algorithms.

## I. Introduction

Over the past decade a variety of machine vision algorithms have emerged. They range from the likes of SIFT[1] and SURF[2], based on interest points and their descriptors to HOG's[3] histograms classified with Support Vector Machines (SVM)[4]. Biologically inspired artificial neural networks have also gained in popularity thanks to the availability of ever increasing computing power required during their learning phase. Convolutional Neural Networks (CNNs)[5] are the current prevailing implementation of artificial neural networks. The efficacy of CNNs is also well known outside the field of machine vision.

Given the numerous practical applications of machine vision it is clear that an efficient implementation of the algorithms above is highly desirable. This paper will revisit the algorithms listed above in view of performing them in the compressed domain (i.e. performing an algorithm directly on a compressed representation of the original data). However, rather than relying on data already compressed in an existing standard (such as JPEG or H.264), a compression framework will be defined as pre-processing step. In other words, the compression step becomes an essential part of the algorithm. The rationale behind this idea is simple: if a compressed signal (image, speech or other) can still be recognized and/or processed by a human being, then it is reasonable to expect that the information lost in the data compression process is not essential to said processing/recognition. Therefore the data compression step can be seen as a filter that eliminates non essential parts of the signal that needs to be processed.

Besides the obvious benefit of processing a smaller amount of information, there are other advantages in working in the compressed domain. Smaller amounts of information mean less bandwidth to an external memory and, hence, power. In specialized hardware it can mean less complex and power hungry circuits.

Of course for this methodology to be beneficial, two conditions must be met:

- The total computational cost of the new algorithm (including the compression step) must be less than the original
- The likely degradation in performance (i.e. a lower recognition rate) due to working in the compressed domain must still be acceptable for a given application

The compression technique used here is based on a fast form of vector quantization called Pyramid Vector Quantization (PVQ). Properties of a quantized vector will then be exploited in order to simplify a variety of algorithms.

## II. Pyramid Vector Quantization

A pyramid vector quantizer[6] (PVQ) is based on the cubic lattice points that lie on the surface of an N-dimensional pyramid. Unlike better known forms of vector quantization that require complex iterative procedures in order to find the optimal quantized vector, it has a simple encoding algorithm.

Given an integer K, any point on the surface of an N-dimensional pyramid $\hat{y}$ is such that

$$\sum_{i=0}^{N-1} |\hat{y}_i| = K \quad (1)$$

with $\hat{y}_i$ integers. The pair of integers N and K, together with (1), completely define the surface of an N-dimensional pyramid indicated here with $P(N,K)$.

In this work a particular type of PVQ will be used, known as product PVQ. Here a vector $\vec{y} \in \mathbb{R}^N$ is approximated by its norm $r = \|\vec{y}\|_2$ (also referred to as "radius" or "length" of the vector) and a direction in N-dimensional space given by the vector that passes between the origin and a point $\hat{y}$ on the surface of the N-dimensional pyramid:

$$r \frac{\hat{y}}{\|\hat{y}\|_2} \quad (2)$$

Note that the direction in N-dimensional space is effectively vector quantized. Null vectors are represented by $r=0$. The radius $r$ can also be quantized with a scalar quantizer. The vector $\hat{y}$ needs to be normalized as it does not lie on the unit hyper-sphere. Given N, increasing K increases the number of quantized directions in N-dimensional space and, hence, the quality of the approximation.

The paper[6] also includes simple algorithms to calculate the number of points $N_p(N,K)$ on the surface of the N-dimensional pyramid. It also provides algorithms to map any point on said surface to an integer $0 \leq i < N_p(N,K)$ and vice-versa. Such mapping provides a much more compact representation of a surface point then a direct bit representation.

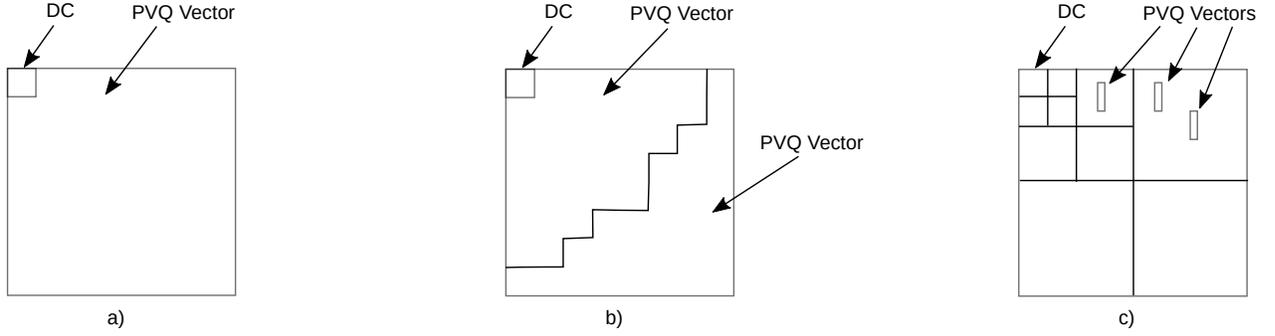

Fig. 1. Possible groupings of AC coefficients into vectors to be PVQ encoded: a single vector in a 8x8 transform a), two vectors in a 8x8 transform b), example of some vectors in a full image wavelet decomposition c).

For example, for N=8 and K=4, each component $\hat{y}_i$ would naively need 4 bits (including the sign), for a total of 8x4=32 bits for the whole vector $\hat{y}$. However, because of the constraint (1), $N_p(8,4)=2816$ and, therefore, less than 12 bits are required to map any $\hat{y} \in P(8,4)$.

The mapping of $\hat{y}$ to an integer is not essential to the vector quantization of $\vec{y}$ but it can be useful in those applications where a quantized vector needs to be stored in a more compact way.

In this work, PVQ is chosen as the pre-processing step proposed above. The properties of PVQ vectors are then exploited to simplify various machine vision algorithms. Also, in the following text, "PVQ encoding" or simply "encoding" a vector $\vec{y}$ will mean finding its closest approximation (2). "Mapping a vector to an integer" will refer to the process that associates a PVQ vector $\hat{y} \in P(N,K)$ to an integer $0 \leq i < N_p(N,K)$. Similarly for its opposite.

The computational cost of PVQ encoding is not very high. The author estimates it to be $O(NlogN)$. In hardware, with a $\log(N)$ number of very simple processing units, encoding at one vector element per clock is easily achievable.

### III. PVQ in Signal Compression

PVQ has been previously used for image compression[7] and CELT[8] audio compression. More recently, also in video compression, as part of Daala[9].

PVQ encoding has been shown in [6] to provide significant mean square error improvements for Laplacian, gamma and Gaussian memoryless sources over the corresponding scalar quantizer. Since images and speech are not modeled well as any of these sources, energy compacting transforms such as DCT, Hadamard transforms or wavelet decompositions are applied before PVQ. After such transforms are applied, all the resulting coefficients (with the exception of the DC) can be modeled as memoryless Laplacian sources.

Details vary, but a simple image compression algorithm that uses PVQ consists of the following steps:
- Divide the image in, say, 8x8 blocks of pixels and apply a transform such as DCT, Hadamard or similar
- Alternatively, perform a wavelet decomposition of the image
- Assemble groups of AC coefficients into one or more N-dimensional vectors. DC coefficients are encoded separately (see Fig. 1)
- PVQ encode said vectors
- The radius $r$ is quantized with a scalar quantizer and the vector $\hat{y}$ is mapped to an integer
- The two resulting numbers per vector are encoded as strings of bits as part of the compressed image

Although described for images, similar algorithms can work on multidimensional signals with multidimensional transforms.

The quality of the compressed image will depend on how finely the radius $r$ was quantized as well as the value of K

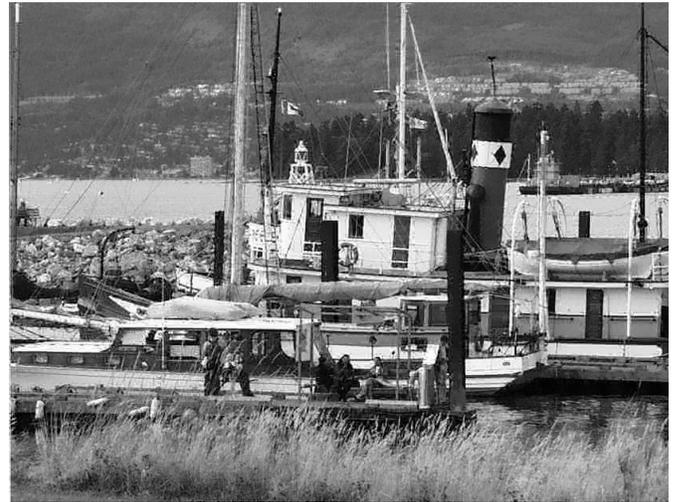

Fig. 2. Example of image divided into 8x8 blocks of pixels, each transformed with the DCT and PVQ encoded to a pyramid P(63,10).

that determines the number of quantized directions in N-dimensional space for the encoded vector.

Fig.2 shows an example of an image divided into 8x8 blocks of pixels, each transformed with the DCT and PVQ encoded to a pyramid P(63,10). The DC coefficient and the radius are not quantized as it is not necessary for most of the algorithms described here.

It is important to note that the compressed image will have a fixed size, without the need for complex bitrate controlling algorithms and/or multiple passes. The downside, compared to variable bitrate algorithms such as JPEG, is that "problem" macroblocks can result in local artifacts because the fixed number of bits that were allocated for that area were insufficient. JPEG suffers from such problems to a lesser extent as it can variably allocate more bits to such blocks.

It is now possible to refine the already mentioned strategy that will be used in this paper:
1. Apply a multi-dimensional energy compacting transform/wavelet decomposition to the signal
2. Weigh different transformed coefficients with different constants. For example, for a DCT transform, this can mean to multiply the transformed coefficients by different constants that enhance or depress particular frequencies
3. Group sets of coefficients together to form one or more vectors and PVQ encode them
4. Perform an algorithm normally performed on the original signal directly on the PVQ vector(s)

Steps 1. and 2. are optional and depend on the nature of the signal.

## IV. Dot Product

We will now look at the dot product between a PVQ vector (2) $\hat{y} \in P(N,K)$ with radius $r$ and an N-dimensional vector $\vec{x} \in \mathbb{R}^N$:

$$r \frac{\hat{y}}{\|\hat{y}\|_2} \cdot \vec{x} = \frac{r}{\|\hat{y}\|_2} \sum_{i=0}^{N-1} \hat{y}_i x_i \quad (3)$$

Since each $\hat{y}_i$ is an integer, each product $\hat{y}_i x_i$ can be expressed as:

$$\hat{y}_i x_i = \begin{cases} 0 & \text{if } \hat{y}_i = 0 \\ \underbrace{x_i + \ldots + x_i + \ldots + x_i}_{|\hat{y}_i| \text{ times}} & \text{if } \hat{y}_i > 0 \\ \underbrace{-x_i + \ldots - x_i + \ldots - x_i}_{|\hat{y}_i| \text{ times}} & \text{if } \hat{y}_i < 0 \end{cases} \quad (4)$$

Equation 4 seems quite obvious but, when considered together with (1), it follows that $\sum_{i=0}^{N-1} \hat{y}_i x_i$ can be calculated with exactly K-1 additions and/or subtractions and no multiplications for each possible $\hat{y} \in P(N,K)$. The result is counter-intuitive but it is the direct consequence of a point $\hat{y}$ belonging to the surface of an hyper-pyramid.

For example, let's consider two vectors $\hat{a} = (0, 0, -3, 0, 1, -1, 0)$, $\hat{b} = (2, 1, 0, 1, 0, 0, 1)$, both belonging to $P(7,5)$ and $\vec{x} = (x_0, \ldots, x_i, \ldots, x_6) \in \mathbb{R}^7$. Then $\hat{a} \cdot \vec{x} = -3x_2 + x_4 - x_5 = -x_2 - x_2 - x_2 + x_4 - x_5$ and $\hat{b} \cdot \vec{x} = 2x_0 + x_1 + x_3 + x_6 = x_0 + x_0 + x_1 + x_3 + x_6$. Both requiring 4 additions/subtractions. This is also true for any

$\hat{y} \in P(7,5)$.

Given that $\sqrt{K} \leq \|\hat{y}\|_2 = \sqrt{\sum_{i=0}^{N-1} \hat{y}_i^2} \leq K$ and that K is quite small for practical purposes, the value $\frac{1}{\|\hat{y}\|_2}$ can be

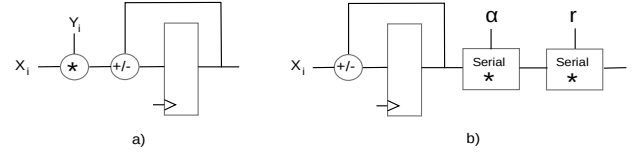

Fig. 3. Serial architecture for dot product a) serial architecture for PVQ dot product b).

pre-calculated in a small look-up table. Therefore the dot product between a PVQ approximated vector (2) and a vector $\vec{x} \in \mathbb{R}^N$ takes K-1 addition/subtractions, one multiplication by the radius $r$ and one by $\frac{1}{\|\hat{y}\|_2}$. Depending on the particular algorithm, a normalized version of the PVQ approximation might be required. In this case $r=1$ and only one multiplication is required. This is to be compared to N multiplications and N-1 additions for the general dot product of N-dimensional vectors. K is also often smaller than N in practice.

It is clear that any algorithm where the dot product operation (or any operation that can be re-conduced to a dot product, like, for example, a linear combination of inputs) is commonly used can take advantage of the smaller computational cost described. The method described also works if the components $x_i$ are not just scalars as in the example but, also, vectors, matrices or tensors.

Of course one has to take into account all the processing required to create the PVQ approximation. Thus the total computational cost is only reduced if the latter can be spread across multiple dot product operations.

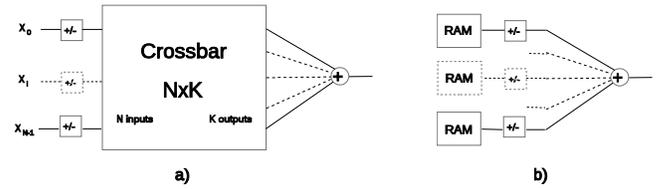

Fig. 4. Parallel architecture for PVQ dot product a) same for FPGA b).

The result given is particularly useful in a hardware implementation. In fact, while specialized DSP instructions can make dot product easier with multiply-accumulating instructions (MAC) in software, being able to use mostly adders can be a substantial advantage in dedicated hardware. An example can be seen in a serial implementation of a dot product digital circuit in Fig. 3b with $\alpha = \frac{1}{\|\hat{y}\|_2}$.

The architecture in Fig. 3a performs a dot product by multiplying to vector elements and accumulating them. It works serially and it takes N cycles for an N-dimensional vector. The architecture in Fig. 3b performs a dot product between a PVQ and an unconstrained vector. It takes K cycles, regardless of the sparsity of the PVQ vector. It works by adding $x_i$ components $\hat{y}_i$ times. It is followed by two serial multipliers. The design is pipelined: while the K-1 additions are performed on vector, the serial multipliers will work on the previous dot product results. If K is large enough, it might be possible to perform two multiplication serially with the same multiplier. Serial multiplication in hardware is effectively an adder and accumulator.

Two parallel architectures for PVQ dot product are shown in Fig. 4. The multiplications by $r$ and $\alpha$ are omitted for clarity. In Fig. 4a the N-dimensional vector $\vec{x}$ is input to the NxK crossbar that selects the elements $x_i$ according to the non zero elements $\hat{y}_i$, $|\hat{y}_i|$ times. For example, for the vectors $\hat{a}$ and $\hat{b}$ mentioned above, the crossbar would select $(x_2, x_2, x_2, x_4, x_5)$ and $(x_0, x_0, x_1, x_3, x_6)$ respectively after sign inversion (if required). The K outputs of the crossbar are added together before the two multiplications. An alternative architecture is proposed in Fig. 4b that is more suitable for FPGAs. Most FPGAs implement small distributed memories very efficiently. A total of K memories are used, each of containing a complete copy of $\vec{x}$. Again, for each non zero $\hat{y}_i$, $x_i$ is selected $|\hat{y}_i|$ times. For $\hat{a}$ and $\hat{b}$ the selection patterns will be $(2,2,2,4,5)$ and $(0,0,1,3,6)$ respectively. The rest of the architecture is identical. The multiple additions can be implemented as a tree of adders. The architecture can be pipelined reaching over 200 million dot products per second in low end FPGAs.

In comparison, a parallel version of the ordinary dot product (not shown) would require N multipliers and N-1 adders, each with a much larger datapath, for a total of many more resources and power.

## V. Support Vector Machines

Support Vector Machines (SVMs) are powerful classifiers that have been successfully used in variety of cases, including image recognition. Detailed description of SVMs is beyond the scope of this article. Suffice to say that a SVM can be trained to signal if a vector $\vec{x} \in \mathbb{R}^N$ belongs or not to a given class if:

$$\vec{w} \cdot \Phi(\vec{x}) + b > 0 \qquad (5)$$

Where $\vec{w}$ and $b$ are constants that result from the training of the SVM and $\Phi(\vec{x})$ is a given vector function applied to the input $\vec{x}$. In case of linear SVM $\Phi(\vec{x}) = \vec{x}$.

The dot product in (5) immediately suggests how the methodology outlined above can be applied here. $\Phi(\vec{x})$ Is PVQ encoded (as usual, an energy compacting transformation might precede this operation if needed) and then the dot product by $\vec{w}$ can be performed with only K-1 additions (assuming $\hat{y} \in P(N,K)$) and two multiplication as explained in section IV. This is particularly useful in case multiple classifiers need to be applied to the same input vector $\vec{x}$ as PVQ encoding is only performed once, before multiple classifications.

As it is advised to normalize the input to SVMs for best results, we can assume $r=1$ and reduce the number of required multiplications by one. In case of the simplest linear SVM with $b=0$ the two multiplications by $r$ and $\frac{1}{\|\hat{y}\|_2}$ are not necessary as they do not affect the sign in (5).

Training of the SVM might be also performed directly on normalized $(r=1)$ PVQ encoded vectors.

## VI. HOG Algorithm

HOG (Histogram of Oriented Gradients) is a popular image recognition algorithm. Details are omitted here but, broadly speaking, the algorithm works by dividing an image into square blocks called cells. For each cell, a histogram of pixel gradient orientations is built. A popular choice for the number of orientations in the histogram is 9. Therefore, in this particular case, each cell will consist of a 9-dimensional vector. Cells are then grouped to form features. A common choice is forming a feature by grouping 2x2 cells, resulting in a 36-dimensional vector which is then normalized. Features overlap and therefore some cells are shared. Once the image is processed into feature, a window is slid over it. All the features contained in the window are used as input vector to one of more SVMs in order to detect an object.

For example, in the original HOG paper, a window of 7x15 features was used to detect pedestrians. Since each feature is a 36-dimensional vector, the total number of inputs to the SVM is 7x15x36=3780 elements wide. This means 3780 multiplications plus 3779 additions for a single detection. This number rapidly balloons for multiple SVMs as the window slides across the image, at different scales and multiple frames per second in real time applications.

In this paper it is proposed to follow all the same steps except for the creation of the features where the normalization step is substituted with PVQ encoding. In particular, the 36-dimensional vector formed by grouping the 4 cells given in the example above is encoded to P(36,K) and normalized $(r=1)$. Preliminary tests from the author seem to indicate that HOG histograms can be modeled as Laplacian sources and, as such, they are suitable for PVQ encoding without additional transformations.

After the image has been processed into this new type of features, the algorithm can proceed as previously described with a sliding window that gathers features to be classified by SVMs. The only difference is that now it is possible to use techniques described in sections IV and V in order to reduce the number of operations.

A comparison can be made with the same 7x15 window of features and the same SVM with a total of 3780 inputs (and weights). After PVQ encoding, each feature in the window will be $\vec{F}_{ij}=\frac{\hat{f}_{ij}}{\|\hat{f}_{ij}\|_2}$ and $\hat{f}_{ij} \in P(36,K)$. The vectors $\vec{w}_{ij}$ will indicate the SVM weights related to each feature $\vec{F}_{ij}$. The indexes $0 \leq i \leq 15$ and $0 \leq j \leq 7$ will indicate the position within the window. Then, assuming a linear SVM:

$$\sum_{i=0}^{14}\sum_{j=0}^{6} \vec{w}_{ij} \cdot \vec{F}_{ij} + b = \sum_{i=0}^{14}\sum_{j=0}^{6} \frac{\vec{w}_{ij} \cdot \hat{f}_{ij}}{\|\hat{f}_{ij}\|_2} + b \quad (6)$$

All the $\vec{w}_{ij} \cdot \hat{f}_{ij}$ can be calculated with K-1 addition each for a total of $105(K-1)$ additions (in a 7x15 window). To this we need to add 105 multiplications to scale each dot product and, finally, another 104 additions to get to the final result.

## VII. CONVOLUTIONS AND CNNs

Convolutions are at the heart of many image processing algorithms, including Convolutional Neural Networks (CNNs). Even modeling the neurons of the mammalian visual cortex involves the convolution with Gabor filters[10]. It is therefore fundamentally important to include this operation in the PVQ framework so far described.

The example given here will be in 2D for blocks of MxM pixels. However, as it will be clear, it can be easily extended to multiple dimensions and even non square matrices.

The convolution of two MxM matrices $C$ and $X$ is here defined as:

$$C * X = \sum_{i=0}^{N-1}\sum_{j=0}^{N-1} c_{ij} x_{ij} \quad (7)$$

The result is a scalar. This operator is clearly linear:

$$C*(aX+bY) = aC*X + bC*Y \quad (8)$$

where Y is a MxM matrix and $a$ and $b$ two scalars.

In image processing, when applied to pixels, the matrix C is also known as kernel or mask. Note also that any MxM matrix $X$ with elements $x_{ij}$ can always be expressed as:

$$X = \begin{pmatrix} x_{00}, \cdots, x_{0M-1} \\ \cdots \\ x_{M-10}, \cdots, x_{M-1M-1} \end{pmatrix} = \sum_{i=0}^{M-1}\sum_{i=0}^{M-1} x_{ij} U_{ij} \quad (9)$$

Where $U_{yx}$ are MxM matrices that contain all zeros except for element $(y,x)$ that is equal to one.

We can now apply a transform $T()$ to the MxM block of pixels X as described in section II. The only additional requirement here is for $T()$ to be linear:

$$Q = T(X) = \begin{pmatrix} q_{00}, \cdots, q_{0M-1} \\ \cdots \\ q_{M-10}, \cdots, q_{M-1M-1} \end{pmatrix} \quad (10)$$

As in section II, we will assume that $T()$ is an energy compacting transform and $q_{00}$ the so called DC coefficient to be left aside. The other $N=M^2-1$ coefficients will be grouped in an N-dimensional vector $\vec{y}$. The order of the grouping is not important. For simplicity, we will assume $y_t = q_{ij}$ with $t=Mi+j+1$, $i=(t+1)/M$ and $j=(t+1) \bmod M$. We can now PVQ encode the vector $\vec{y}$ resulting in the approximation (2). The approximated matrix $\hat{Q}$ will be :

$$\hat{Q} = \begin{pmatrix} q_{00}, \alpha \hat{y}_0, \cdots \\ \cdots \\ \alpha \hat{y}_t, \cdots, \alpha \hat{y}_{N-1} \end{pmatrix} = q_{00} U_{00} + \alpha \sum_{t=0}^{N-1} \hat{y}_t U_{ij} \quad (11)$$

With $\alpha = \frac{r}{\|\hat{y}\|_2}$ and $\hat{y} \in P(N,K)$. We can now apply the inverse transform $T^{-1}()$ to the approximated matrix $\hat{Q}$, creating a matrix of pixels $\hat{X}$ and then perform a convolution product with $C$, effectively calculating an approximation of (7):

$$C*(T^{-1}(\hat{Q})) = q_{00} C*(T^{-1}(U_{00})) + \alpha \sum_{t=0}^{N-1} \hat{y}_t C*(T^{-1}(U_{ij})) \quad (12)$$

Equation (12) follows directly from the linearity of the transform and that of the convolution product. Note also that the terms $C*(T^{-1}(U_{ij}))$ are scalars that can be pre-calculated and stored in an array of size $M^2$, just like the coefficients of $C$. More importantly, the second term of (12) is a dot product with a PVQ vector and, as such, it can be calculated with K-1 additions and 2 multiplications. Another addition and multiplication is needed for the DC term.

Therefore, once a block of pixels is transformed and PVQ encoded, multiple convolutions can be performed on the same block with low computational cost. This can be very useful in large CNNs where the first layer consists of many convolutions performed on the same block or where a large battery of Gabor filters is applied at each step.

Another way to approximate (7), if the transform $T()$ is orthogonal (such as DCT, Hadamard, Haar wavelet), is to remember that:

$$C*X = \sum_{i=0}^{M-1}\sum_{j=0}^{M-1} c_{ij}x_{ij} = T(C)*T(X) = \sum_{i=0}^{M-1}\sum_{j=0}^{M-1} w_{ij}q_{ij}$$
(13)

Equation (13) is true up to a constant and $w_{ij}$ are the elements of the matrix $T(C)$. This suggests an alternative to (12) to approximate (7):

$$C*(T^{-1}(\hat{Q})) = q_{00}w_{00} + \alpha \sum_{t=0}^{N-1} \hat{y}_t w_{ij}$$
(14)

Again, note the dot product with the PVQ vector $\hat{y}$ in the second term of (14).

It is worth remembering the these results, while given for the 2D case, they can be easily extended to multi-dimensional transforms.

Although more testing is necessary, there are good reasons to expect this methodology to work well for at least the first layer of a CNN applied to images and speech recognition. In fact, all the elements of the image in Fig. 2 are clearly recognizable and there are no practical differences with a moderately JPEG compressed image. K is only equal to 10 and that means that any convolution with a kernel smaller or equal in size to 8x8 can be performed with only 9 additions and two multiplications. Moreover, at least in principle, there is no difference between performing a convolution in the pixel domain on the reconstructed image from PVQ vectors or directly on its PVQ representation with (12) or (14): the result will be the same.

Even though it looks like this approach should work on the first layer of a CNN, it remains to be seen if it can be applied to subsequent layers. Much will depend on the type of "images" CNNs produce deeper in the network.

Finally, the same approach that has lead to (12) and (14) can be applied to the cases in Fig. 1b and Fig. 1C where multiple PVQ vectors are involved.

## VIII. A PVQ Convolution Processor

The previous section has shown how to take a block of pixels, transform it, PVQ encode it and then perform one or more convolutions on the same block using the properties of the dot product with a PVQ vector.

Although the cost of these operations might be tolerated when many convolutions need to be performed on the same block of pixel (as it happens in CNNs), it would be desirable to make this process more efficient.

Convolution kernels in CNNs often overlap and this suggests that the same PVQ encoded block can be re-used for convolutions with kernels that are only partially overlapping.

Re-using PVQ vectors implies a buffer to store them.

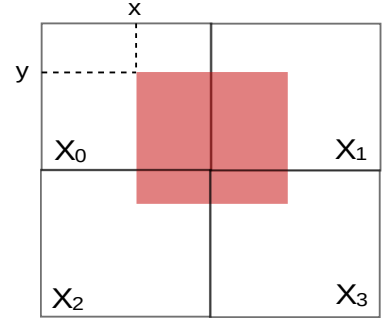

Fig. 5. Convolution kernel C (in red) partially overlapping blocks $X_1, X_2, X_3, X_4$.

Fig. 5 shows a kernel C (in red) $M_1 x M_1$ overlapping four blocks of pixels $X_1, X_2, X_3$ and $X_4$ $M x M$ pixels each. For simplicity, we will assume that C never overlaps more than four blocks: this can be easily extended to any other case. Then we have:

$$C*X = C_0*X_0 + C_1*X_1 + C_2*X_2 + C_3*X_3$$
(15)

Where X is the block of pixels perfectly overlapping with C and $C_{0-3}$ are the kernels perfectly overlapping with the $X_{0-3}$ block. The $C_{0-3}$ kernels coefficients are all zeros, except for the portion in red which coincides with C coefficients.

Equation (15) can be calculated approximately with either (12) or (13). Both equations need the $C_{0-3}$ transformed.

The kernels $C_{0-3}$ will depend on the shift (x,y) from the origin (see Fig. 5). In many CNNs, depending on the kernel size and the stride, only a few pairs of shifts (x,y) might be required. This means that all the transformed $C_{0-3}$ can be pre-calculated and stored. Alternatively, they can be created on the fly.

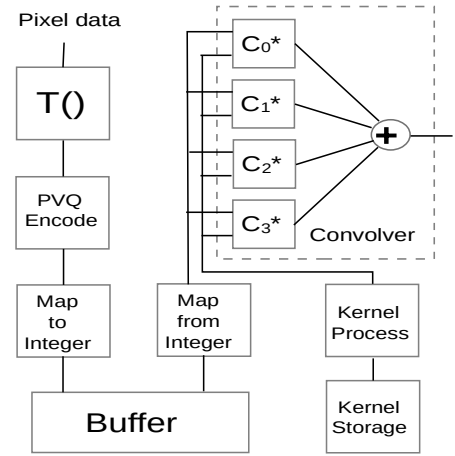

Fig. 6. A PVQ convolution processor.

These ideas together form the blueprint for the architecture of a PVQ convolution processor shown in Fig. 6. Here an image is subdivided in non-overlapping blocks of pixels. Each of them is transformed with a linear (and possibly orthogonal) transform. PVQ encoding follows and, optionally, mapping to an integer. This information, together with the DC component of the block and the radius of the PVQ vector, are stored in a buffer in a very compact way, 4-8 times smaller than their pixel counterpart. The purpose of the buffer is to allow the re-use of PVQ encoded blocks without repeating the encoding. Four PVQ encoded blocks are then fetched from the buffer, mapped back to a vector if necessary, and sent to the convolver.

The convolver uses (15) to apply the kernel C over four PVQ encoded blocks. Each of its elements is based on (12) or (14) which, in turn, can be implemented with architectures shown in Fig. 4 (plus a multiplier and an adder for the DC component). As discussed, the transformed kernel coefficients for $C_{0-3}$ can be pre-calculated and stored in a memory (the kernel storage in this case) or calculated starting from the original C, according to the shift (x,y) shown in Fig. 5.

Whatever the case, if this architecture is to operate at one convolution per clock cycle, the convolver needs to be fed four PVQ encoded blocks as well as all the coefficients for $C_{0-3}$. The former is pretty simple, thanks to the buffer and the low bandwidth of PVQ encoded vectors. The latter is not as easy, due to the large number of $C_{0-3}$ coefficients. Again, the buffering of PVQ vectors can help. In fact it can hold the $C_{0-3}$ kernels constant while calculating convolutions with the PVQ blocks in the buffer that share the same (x,y) shift. In other words, calculate the convolutions only for the blocks that share the same overlapping pattern with the kernel C. In the meantime, new coefficients for the next group of $C_{0-3}$ kernels coefficients are loaded in a shadow register for the architecture in Fig. 4a) or in another memory bank for the one in Fig.4b), without interrupting the convolutions to update the coefficients. This way the transfer of the kernels' coefficients can happen with low bandwidth. Once the transfer is finished, the new coefficients can be immediately active with a register transfer for a) or a RAM bank switch for b) without any lost clock cycles.

In any case the architecture shown in Fig. 6 is just a blueprint and many improvements and changes are possible. For example, multiple convolvers can work in parallel by simply sharing the buffer.

## IX. A Non-PVQ Interlude

Although this paper is about using PVQ in the field of machine vision, a careful reader will have noticed that many of the ideas showed in the previous two sections still apply if PVQ is not used. PVQ is a form of quantization and, if we used scalar quantization after the transformation of a block of pixels, equations similar to (12) and (14) could still be derived.

Equation (15) would still be true as well as many of the concepts in section VIII.

One problem with scalar quantization is that, after it has been applied to a transformed block of pixels, it will result in an unpredictable number of non-zero coefficients. In order to avoid this problem, we will define a simpler way to approximate a transformed block of MxM pixels by keeping $K < M^2$ coefficients with the largest absolute value and setting all the others to zero. This can still be seen as a form of quantization.

In this case, (13) will still be true but, in rightmost part, only K terms will be non zero. It follows that an approximated convolution can be calculated with only K multiplications and K-1 additions. The steps of the algorithm will be as follows:

- Apply an orthogonal transform to a block of MxM pixels
- Retain only K coefficients with the largest absolute value and set all others to zero
- Now the convolution with a transformed kernel can be performed using (13) with only K multiplications and K-1 additions

If the transform is only linear, than the algorithm can still be used with coefficients $C * (T^{-1}(U_{ij}))$.

It is a well known property of the Discrete Fourier Transform (DFT) (and its faster counterpart, the FFT) that convolutions can be performed by simple multiplication in the transformed domain:

$$C * I = DFT^{-1}(DFT(C) DFT(I))$$
(16)

Where $I$ is an image and $C$ is a kernel to be convolved with $I$. Details like padding etc. are here omitted.

Therefore a similar algorithm also works for DFT/FFT:

- Perform a 2D FFT on an image
- Retain only K coefficients with the largest magnitude value and set all others to zero
- Now the convolution with a transformed kernel can be performed using (16) with only K multiplications. Multiple convolutions can be applied.
- Perform the inverse DFT on each convolution result

Since eliminating 90% or more of the lowest magnitude coefficients from a Fourier transformed image still produces a perfectly recognizable image, it is reasonable to expect that a CNN should still work on such images. Therefore the algorithm above is certainly a possibility to reduce the computational load. Note also that there is no reason here to limit oneself to images: similar approaches will work on multidimensional signals.

## X. Vector Cosine Similarity

Given two N-dimensional vectors $\vec{x}, \vec{y} \in \mathbb{R}^N$, the cosine of the angle $\varphi$ between them is:

$$\cos(\varphi) = \frac{\vec{x} \cdot \vec{y}}{\|\vec{x}\|_2 \|\vec{y}\|_2}$$
(17)

If we discount the magnitude of the vectors, (17) can be used as a measure of similarity of the two vectors. In fact, as the vectors tend to the same direction, the angle $\varphi$ tends to 0, $\cos(\varphi)$ tends to 1. Also, if $\vec{x}, \vec{y}$ are normalized, the following is true :

$$\|\vec{x}-\vec{y}\|_2 = \sqrt{2-2\cos(\varphi)} \tag{18}$$

The latter re-enforces the fact that (17) can be used as a measure of similarity when the vectors are normalized.

Let us consider now the cosine of the angle between an N-dimensional vector $\vec{x} \in \mathbb{R}^N$ and a normalized, PVQ encoded vector. The latter is given in (2) with $r=1$, giving:

$$\cos(\varphi) = \frac{\vec{x} \cdot \hat{y}}{\|\vec{x}\|_2 \|\hat{y}\|_2} \tag{19}$$

with $\hat{y} \in P(N,K)$. Equation (19) contains a dot product with a PVQ vector and, as usual, this simplifies its computation.

Let us consider now the following problem: we have a set of vectors $S \subseteq P(N,K)$ and, for a given vector $\vec{x} \in \mathbb{R}^N$, we want to find $\hat{y} \in S$ that is "closest" to the normalized version of $\vec{x}$. This can be found by maximizing (19) over $S$:

$$\underset{\hat{y} \in S}{\operatorname{argmax}} \frac{\vec{x} \cdot \hat{y}}{\|\vec{x}\|_2 \|\hat{y}\|_2} = \underset{\hat{y} \in S}{\operatorname{argmax}} \frac{\vec{x} \cdot \hat{y}}{\|\hat{y}\|_2} \tag{20}$$

Which can be calculated with only K-1 additions and one multiplication for each $\hat{y} \in S$. It is clear that this problem is akin to a classification problem where, given a new vector, we try to find the most similar to a group of known vectors.

The framework described has also relevance in another area of machine vision: key or interest points, the realm of algorithms such as SIFT and SURF. Again, description of these algorithms is beyond the scope of this paper. Here we will remember that both consist of essentially three parts:

1. An image is scanned for "fixed" points with detectors that are hopefully invariants to illumination, scale or rotational changes.
2. A small area around a fixed point is used to create a "point descriptor". The descriptor is crafted to be resilient to image changes.
3. Point descriptors can be matched to other point descriptors for a variety of machine vision tasks such as image recognition and/or Simultaneous Localization And Mapping (SLAM)

It is proposed here to use PVQ encoding for the creation of point descriptors. The reason is the capability of PVQ encoding to represent pixel data in a compact, fixed size as well as the ease to compare the similarity of different descriptors as shown by (19). PVQ encoding also acts as a dimensionality reduction technique that automatically allocates more bits to the largest components of a vector within the total bit budget determined by the parameter K.

For example, the SIFT and GLOH[11] descriptors are based on histograms of gradients seem to be particularly suited to PVQ encoding that could substitute normalization. GLOH also requires its dimensions reduced to 64 with Principal Component Analysis (PCA) and that could be avoided by PVQ encoding.

Another important property of PVQ encoding, as already mentioned in section II, is the possibility of mapping each encoded vector to an integer that needs less bits than a naive representation of the same vector. This is very useful in case many descriptors need to be stored in an external memory as it would reduce the amount of memory used as well as bandwidth and power.

Potentially, any descriptor that requires a normalization and/or dimensionality reduction is a candidate to PVQ encoding.

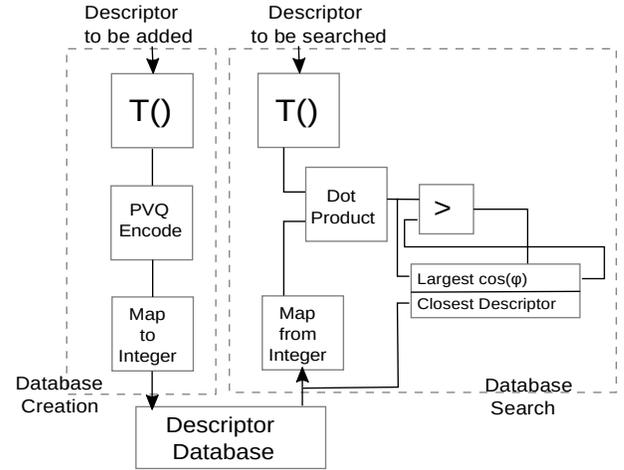

Fig. 7. A PVQ keypoint processor.

Fig. 7 shows the possible architecture for a keypoint processor. It consists of two separate sections or processes. In the database creation section, new descriptors are processed and added to a database. This can be done off-line where a database is pre-loaded with key points (for example known landscapes for SLAM) or dynamically where new descriptors are added as they are encountered. The steps have already been described before, including an optional transform that might be required to improve the statistics of the descriptor's components for PVQ encoding. As the descriptors are normalized, the radius of the PVQ vector is discarded.

During the search phase, a new descriptor is compared to PVQ descriptors in the database, effectively performing the operation in (20). This is done by performing a dot product and comparing the result with the content of a register (initially set to zero). If the former is larger than the latter, the value is stored together with the identification of the closest vector found so far.

## XI. CONCLUSION AND FUTURE WORK

This paper proposes to use PVQ to speed up some well known machine vision algorithms. Future work will include testing of the ideas presented and, if successful, the design of specialized hardware demonstrating the technology.

Particularly interesting is the question of whether this methodology can be applied to simplify the computation of deeper layers of CNNs.

However, it is clear that any algorithm that makes use of dot products, especially on sparse data, could take advantage of this framework and the use of PVQ encoding should be investigated. This could include algorithms in fields unrelated to machine vision such as Natural Language Processing which makes use of cosine vector similarity.

## References

[1] D. G. Lowe, "Distinctive Image Features from Scale-Invariant Keypoints" International Journal of Computer Vision 60(2), 91-110, 2004.

[2] H. Bay, A. Ess, T. Tuytelaars, L. Van Gool "Speeded-Up Robust Features (SURF)" Computer Vision and Image Understanding Volume 110 Issue 3, June, 2008.

[3] N. Dalal, B. Triggs, "Histograms of Oriented Gradients for Human Detection", Conference on Computer Vision and Pattern Recognition (CVPR), 2005

[4] https://en.wikipedia.org/wiki/Support_vector_machine

[5] Y. LeCun, L. Bottou, Y. Bengio, P. Haffner, "Gradient-Based Learning Applied to Document Recognition" *Proceedings of the IEEE 86 (11): 2278–2324, 1998*.

[6] T. R. Fischer, "A Pyramid Vector Quantizer" IEEE Transactions on Information Theory, Vol. IT-32, No. 4, July 1986.

[7] T. H. Meng, B. M. gordon, E. Tsern, A. C. Hung, "Portable Video-on-Demand in Wireless Communication", Proceedings of the IEEE, VOL 83, NO. 4, April 1985.

[8] J. Valin, T. B. Terriberry, C. Montgomery, G. Maxwell, "A High-Quality Speech and Audio Codec With Less Than 10 ms Delay" (17 April 2009), IEEE Signal Processing Society, ed.

[9] J.-M. Valin, T. B. Terriberry, "Perceptual Vector Quantization for Video Coding", Proceedings of SPIE Visual Information Processing and Communication Conference, 2015.

[10] T. Serre, L. Wolf, S. Bilenschi, M. Reisenhuber, T. Poggio. "Robust ObjectRecognitionwith Cortex-Like Mechanism", IEEE Transactionson Pattern Analysis and Machine Intelligence, VOL. 29, NO. 3, March 2007.

[11] K. Mikolajczyk, C. Schmid "A performance evaluation of local descriptors", IEEE Transactions on Pattern Analysis and Machine Intelligence, 10, 27, pp 1615--1630, 2005.